\documentclass[10pt,twocolumn,letterpaper]{article}

\usepackage{cvpr}              
\usepackage{amssymb}
\usepackage[accsupp]{axessibility}
\usepackage{multirow}
\usepackage{diagbox}
\usepackage{floatrow}
\usepackage{graphicx}
\usepackage[dvipsnames]{xcolor}

\definecolor{cvprblue}{rgb}{0.21,0.49,0.74}
\usepackage[pagebackref,breaklinks,colorlinks,citecolor=cvprblue]{hyperref}
\def\paperID{2} 
\def\confName{CVPR}
\def\confYear{2024}

\title{FloCoDe: Unbiased Dynamic Scene Graph Generation with Temporal Consistency and Correlation Debiasing}

\author{Anant Khandelwal\\
Glance AI\\
{\tt\small anant.iitd.2085@gmail.com}
}
\begin{document}
\maketitle
\begin{abstract}
Dynamic scene graph generation (SGG) from videos requires not only a comprehensive understanding of objects across scenes but also a method to capture the temporal motions and interactions with different objects. Moreover, the long-tailed distribution of visual relationships is a crucial bottleneck for most dynamic SGG methods. This is because many of them focus on capturing spatio-temporal context using complex architectures, leading to the generation of biased scene graphs. To address these challenges, we propose \textsc{FloCoDe}: \textbf{Flo}w-aware Temporal Consistency and \textbf{Co}rrelation \textbf{De}biasing with uncertainty attenuation for unbiased dynamic scene graphs. \textsc{FloCoDe} employs feature warping using flow to detect temporally consistent objects across frames. To address the long-tail issue of visual relationships, we propose correlation debiasing and a label correlation-based loss to learn unbiased relation representations for long-tailed classes. Specifically, we propose to incorporate label correlations using contrastive loss to capture commonly co-occurring relations, which aids in learning robust representations for long-tailed classes. Further, we adopt the uncertainty attenuation-based classifier framework to handle noisy annotations in the SGG data. Extensive experimental evaluation shows a performance gain as high as 4.1\%, demonstrating the superiority of generating more unbiased scene graphs.
\end{abstract}
\vspace{-4mm}    
\vspace{-2mm}
\section{Introduction}
\label{sec:intro}
Scene graph generation for videos (VidSGG) aims to represent the video in the form of a dynamic graph that is able to capture the temporal evolution of the relationships between pairs of objects. VidSGG has direct applications for various downstream tasks, such as visual question answering \cite{antol2015vqa, tapaswi2016movieqa, xiao2021next}, video captioning \cite{xu2015show}, and video retrieval \cite{dong2021dual, snoek2009concept, wei2019neural}, etc. VidSGG is considered more challenging compared to its image-based counterpart, as the relations between identified object pairs are dynamic along the temporal dimension, making it a multi-label problem. At the current stage, VidSGG is still in its nascent phase compared to SGG (scene graph generation) for static images \cite{desai2021learning, krishna2017visual, li2022sgtr, li2017scene, lin2020gps, lu2016visual, tang2019learning, zellers2018neural, zhang2019graphical}.

Several works \cite{cong2021spatial, liu2020beyond, qian2019video, teng2021target, goyal122021simple, ji2021detecting, li2022dynamic} have proposed solutions for VidSGG, mostly employing spatio-temporal sequence processing with transformers \cite{arnab2021vivit, carion2020end, han2022survey, ju2022prompting, nawhal2021activity, vaswani2017attention, sun2019videobert}. However, many of these methods focus on building complex models to effectively aggregate spatio-temporal information in a video, without adequately addressing data imbalance in the relation/predicate classes. While their performance is often good in terms of the Recall@K metric, which is biased towards frequent classes, a more comprehensive metric, mean-Recall@K, has been proposed \cite{chen2019knowledge, tang2019learning} to assess performance in the presence of low-frequency classes, providing an overall view of SGG models rather than considering only high-frequency classes.

Although recent methods \cite{nag2023unbiased, pu2023spatial} have proposed addressing class imbalance using memory-based debiasing and uncertainty attenuation for classification, the memory debiasing approach is based on learnable attention weights, which carries a risk of bias towards high-frequency classes. A case of this issue is illustrated in Figure \ref{qual} during a qualitative comparison with our method. To overcome these limitations, some works \cite{li2021interventional, xu2022meta} have attempted to address biased relation predictions. \cite{li2021interventional} proposed weakening the false correlation between input data and predicate labels, while \cite{xu2022meta} considered biases in a meta-learning paradigm. Although these approaches mitigate the long-tail problem to some extent, the performance is still not satisfactory.

Our analysis (see Table \ref{ablation}) pinpoints a significant issue in existing dynamic Scene Graph Generation (SGG): the inaccurate detection of objects across video frames. Owing to this issue, we propose leveraging flow-warped features in the temporal dimension to handle dynamic fluctuations in videos. Unlike previous methods \cite{nag2023unbiased} that employ learnable read from memory, we propose debiasing predicate embeddings during the generation stage. This is achieved by ensuring an unbiased correlation between predicates and objects. Additionally, existing methods \cite{nag2023unbiased} overlook label correlations, which can provide valuable clues during relation classification, especially for long-tailed predicate classes. We propose supervised multi-label contrastive loss to consider label correlations. This aims to bring together predicate representations with overlapping classes and push apart negative samples without shared classes. Further, we introduce a Mixture of Logit Networks (MLNs) to calculate aleatoric (data) and epistemic (model) uncertainty\footnote{Aleatoric uncertainty pertains to the intrinsic and unavoidable aspects of uncertainty within the data generation process, such as measurement noise. Conversely, epistemic uncertainty encompasses the uncertainty associated with the model itself, and this type of uncertainty may diminish with an increase in the amount of training data available.}. We propose an uncertainty-aware regularization method aimed at improving the capacity to comprehend inconsistent output patterns, spanning both clean and noisy distributions. This method not only enables robust learning of the fundamental target distribution but also skillfully captures the injected noise patterns inherent in the data generation process. 
Combining these elements, our framework, \textsc{FloCoDe} (\textbf{Flo}w-aware Temporal Consistency and \textbf{Co}rrelation \textbf{De}biasing with Uncertainty Attenuation), makes significant contributions: 1) We propose modeling both aleatoric and epistemic uncertainty and label correlations for more unbiased scene graphs. 2) Introducing a novel correlation-guided debiased learning of predicate embeddings. 3) Incorporating flow-aware, temporally consistent object detection for precise node classification in scene graphs. 4) Achieving substantial improvements in mR@K \cite{tang2019learning} and R@K, showcasing its superiority in generating unbiased scene graphs.

\section{Related Work}
\label{sec:related}
\textbf{Image Scene Graph Generation}: Image-based scene graph generation (ImgSGG) refers to the task of generating a structured graph summary that represents objects as nodes and their relationships (formally known as predicates) as edges within an image. Extensive efforts have been dedicated to ImgSGG, particularly in comparison to the Visual Genome (VG) benchmark \cite{krishna2017visual}. Previous research has focused on developing efficient approaches for aggregating spatial context \cite{li2017scene, lin2020gps, lu2016visual, zellers2018neural, zhang2019graphical}. Concurrently, recent studies \cite{tang2020unbiased, tang2019learning, wen2020unbiased} have concentrated on addressing foundational challenges, such as mitigating biased scene graphs resulting from long-tailed predicate distributions and dealing with noisy annotations in datasets.\\
\textbf{Video Scene Graph Generation (VidSGG)}: Researchers have made significant strides in understanding spatial context within images, prompting an exploration of spatial context and temporal correlations between objects in videos. In the realm of video scene graph generation (VidSGG), challenges akin to those in image-based scene graph generation persist, including long-tailed predicates and noisy annotations. Action Genome \cite{ji2020action}, a prominent VidSGG benchmark, introduces an additional hurdle of addressing temporal fluctuations across frames. Various approaches \cite{qian2019video, zheng2022vrdformer, tsai2019video, liu2020beyond, teng2021target} have employed object-tracking mechanisms to manage temporal fluctuations, but these models often come with high computational costs, memory consumption, and performance issues due to information accumulation from irrelevant frames. STTran \cite{cong2021spatial} offers a robust baseline, using a spatial encoder and a temporal decoder to implicitly extract spatial-temporal contexts. Other works \cite{li2022dynamic, wang2022dynamic} focus on extracting temporal correlations, ensuring temporal continuity through co-occurrence patterns or proposing pre-training paradigms. Building on the success of transformer-based models \cite{arnab2021vivit, nawhal2021activity, sun2019videobert, vaswani2017attention}, which excel in sequence processing for spatio-temporal context, there remains a challenge of bias towards high-frequency classes. TEMPURA \cite{nag2023unbiased} addresses long-tail bias with an uncertainty-guided loss function. In our contribution, we extend this by exploring label correlation to further enhance predictive accuracy in VidSGG.
\section{Method}
\label{sec:dynsgg}
\textbf{Problem Statement}: Given a video $\mathcal{V} = \{I_1, I_2, I_3,..., I_T\}$, the goal of dynamic SGG is to generate scene graphs denoted as $\mathcal{G} = \{G_t\}_{t=1}^{T}$ of video $\mathcal{V}$ consisting of $T$ frames. $G_t = \{V_t, E_t\}$ is the scene graph of frame $I_t$, where $V_t$ is the set of nodes and $E_t$ is the set of edges representing relations between nodes in $V_t$. Nodes in $V_t$ are connected to each other using predicates in $E_t$, forming multiple $<$\textit{subject-predicate-object}$>$ triplets. The sets of object and predicate classes are referred to as $\mathcal{Y}_o = \{{y_{o}}_{1}, {y_{o}}_{2}, {y_{o}}_{3},..., {y_{o}}_{c_o}\}$ and $\mathcal{Y}_r = \{{y_{r}}_{1}, {y_{r}}_{2}, {y_{r}}_{3},..., {y_{r}}_{c_r}\}$ respectively.\\
\textbf{Object Detection and Relation Representation}: With the use of an off-the-shelf object detector (Faster-RCNN \cite{ren2015faster}), we obtain the set of objects $O_t = \{o_{i}^{t}\}_{i=1}^{N(t)}$, where $N(t)$ is the number of objects detected in frame $I_t$. Each object in a $t^{th}$ frame is denoted as $o_{i}^{t} = \{b_{i}^{t}, v_{i}^{t}, c_{o_i}^{t}\} $ where $b_{i}^{t} \in \mathbb{R}^4$ being the bounding box, $v_{i}^{t} \in \mathbb{R}^{2048}$ the RoIAligned \cite{he2017mask} proposal feature of $o_{i}^{t}$ and $c_{o_i}^{t}$ is its predicted class. However, the object class $c_{o_i}^{t}$ fluctuates across the frames and is not coherent even for the same object. Existing works \cite{teng2021target} address this by incorporating object tracking algorithms; opposed to this, our strategy compensates for these dynamic fluctuations using flow-warped features and ensures temporal coherence. Specifically, we extracted the base features $f_t, t \in [1, T]$ and ROIs from Faster R-CNN using ResNet-101 \cite{he2016deep}. We warp these base features using the temporal flow and compute the RoIAligned warped object features $v_{i}^{t \rightarrow t'}$ ($t'$ represents the immediate previous frame that contains the $i^{th}$ object) using the predicted ROIs.
\subsection{Temporal Flow-Aware Object Detection}
\label{objd}
Object detectors trained on static images are prone to misclassify the same object in different frames. Existing methods \cite{cong2021spatial, ji2021detecting, li2022dynamic, wang2022dynamic} either use $c_{o_i}^{t}$ obtained from object detection in each frame or use object features to classify objects. However, these methods do not compensate for temporal fluctuations in the videos. Inspired by FGFA\cite{zhu2017flow}, which uses flow-guided feature aggregation for object detection in videos, we propose to leverage flow-warped features and temporal processing for consistent object detection across frames.
\begin{figure*}
    \centering
    \includegraphics[width=0.8\textwidth]{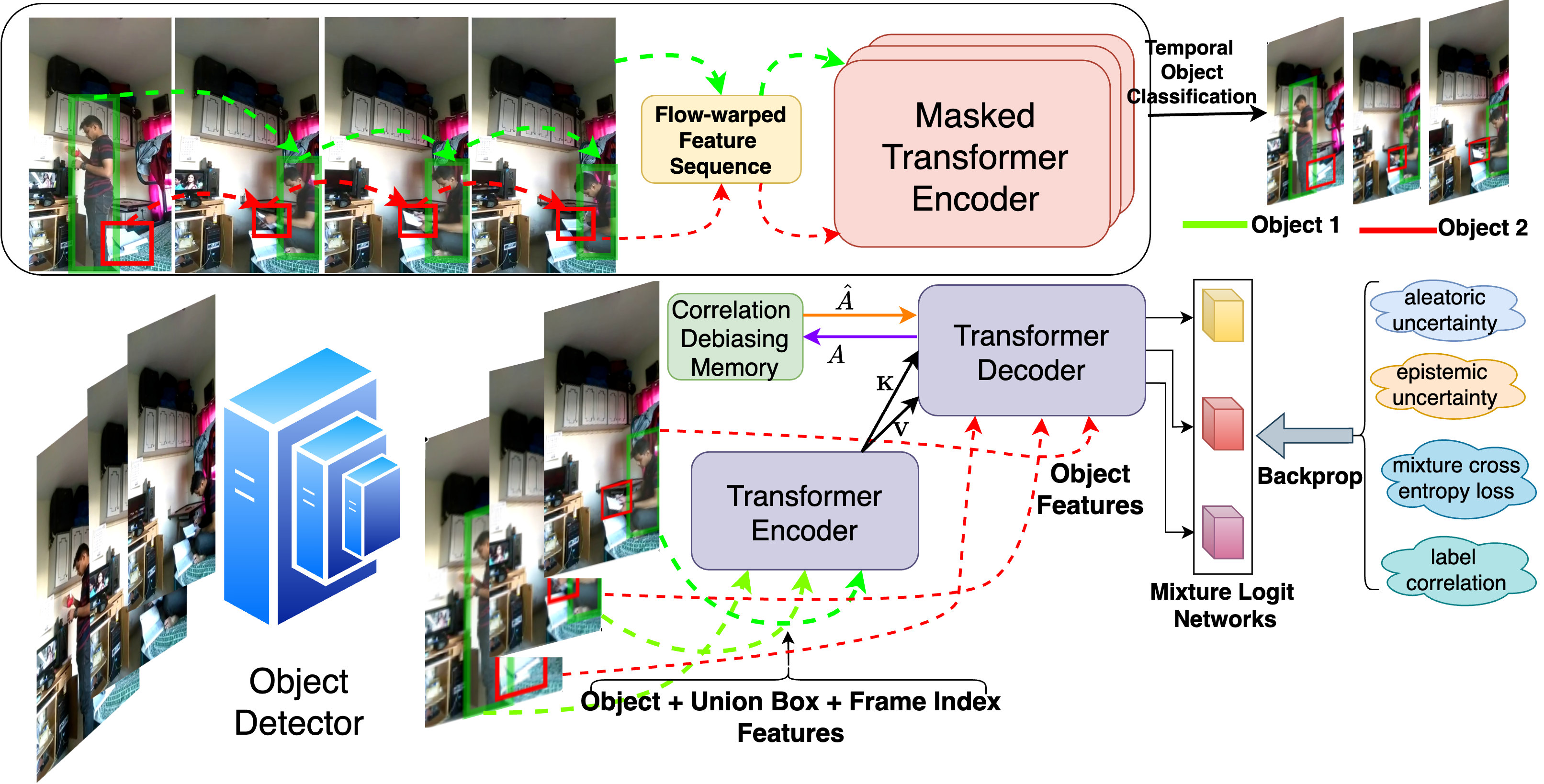}
    \caption{\textsc{FloCoDe}: Each RGB frame is passed to object detector, and the corresponding object proposals are passed to temporal flow-aware object classification. Finally, we predict relations (with correlation debiasing, Mixture Logit Network for uncertainty estimation, mixture loss and label correlation loss) from relation representation at the output of transformer decoder to generate unbiased scene graphs}
    \label{fig:enter-label}
    \vspace{-5mm}
\end{figure*}
We introduce \textit{Temporal Flow-Aware Object Detection (TFoD)}, which utilises transformer encoder \cite{vaswani2017attention} with masked self-attention (\textit{TEnc}) to process the set of temporal object sequences $\mathcal{O}_{\mathcal{V}}$, which is constructed as follows:
\begin{multline}
    \mathcal{O}_{\mathcal{V}} = \{\mathcal{O}_{t_1, k_1}^{1}, \mathcal{O}_{t_2, k_2}^{2}, ..., \mathcal{O}_{t_{\hat{C}_o}, k_{\hat{C}_o}}^{\hat{C}_o} \}, \textrm{ where } \\ \mathcal{O}_{t_j, k_j}^{j} = \{v_i^{t_j}, v_i^{{t_j}+1},......, v_{i}^{k_j}\}
\vspace{-4mm}
\end{multline}   
Each entry of $\mathcal{O}_{t_j, k_j}^{j}$ corresponds to an object of the same detected class ${c_{o}}_{j}$, here $1 \leq t_j, k_j \leq T$ and $\hat{\mathcal{C}}_o \leq \mathcal{C}_o$ denoting all the detected classes in a video $\mathcal{V}$. However, the detected class labels can be noisy since they are based only on frame-level predictions; hence, we use the flow-warped feature $v_{i}^{t \rightarrow t'}$ instead of $v_i^{t}$ before feeding it to the transformer encoder. The flow-warped feature is computed as:
\begin{equation}
    f_{t \rightarrow t'} = \mathcal{W}(f_t, \mathcal{F}(I_{t'}, I_{t}))
\end{equation}
where $\mathcal{W}$ is a bilinear warping function \cite{zhu2017deep, zhu2017flow} applied on all the locations for each channel in the feature maps and flow field $\mathcal{F}(I_{t'}, I_{t}) $ is computed from the pre-trained Flow-Net \cite{dosovitskiy2015flownet}, where $t'$ is the index of the immediate frame previous to the $t^{th}$ frame having the same object as depicted in the object sequences $\mathcal{O}_{\mathcal{V}}$. Using $f_{t \rightarrow t'}$ and predicted ROIs, we obtained warped RoIAligned feature, $\mathcal{O}_{t_j, k_j}^{j}$ is given as:
\begin{equation}
   \mathcal{O}_{t_j, k_j}^{j} = \{v_i^{t_j}, v_i^{t_j+1 \rightarrow t_j},......, v_{i}^{k_j}\}
\end{equation}
Each of $\mathcal{O}_{t_j, k_j}^{j}$ is zero-padded to prepare functional input tensor. \textit{Tenc} uses masked multi-head self-attention instead of multi-head self-attention in transformer encoder \cite{vaswani2017attention}. Mask is introduced to learn the temporal dependencies in a unidirectional manner so that the object at frame index $t$ can only attend to objects in previous frames. For any input $\mathbf{X}$, the single head masked attention $\mathbb{A}$ is given as:
\begin{equation}
    \mathbb{A}(\mathbf{Q, K, V}) = softmax \left (  \frac{mask(\mathbf{Q}\mathbf{K}^T)}{\sqrt{D_K}}  \right ) \mathbf{V}
\end{equation}
where $D_K$ is the dimension of $\mathbf{K}$, and $\mathbf{Q, K, V}$ are the query, key, and value vectors, respectively. Here, $\mathbf{Q=K=V=X}$, the multi-head attention is $concat(a_1, a_2,...a_H)W_H$, where $a_i = \mathbb{A}(\mathbf{X}{W_Q}_{i}, \mathbf{X}{W_K}_{i}, \mathbf{X}{W_V}_{i}) $ where ${W_Q}_{i}, {W_K}_{i}, {W_V}_{i} \textrm{ and } W_H$ are the learnable weight matrices. The rest of the components, like residual connection, normalisation, and FFN (feed-forward network), remain the same as in the transformer encoder \cite{vaswani2017attention}. The output of n-layered \textit{TEnc} is given as:
\begin{equation}
    X_{out}^{(n)} = TEnc(X_{out}^{(n-1)}), \textrm{ 
 } X_{out}^{(0)} = \hat{\mathcal{O}}_{V}
\end{equation}
where $\hat{\mathcal{O}}_V = \mathcal{O}_{V} + P_{o}^T$, where $P_{o}^T$ are the fixed positional embeddings injecting the temporal position of objects. Inspired by the properties of neural collapse \cite{papyan2020prevalence} we prefixed the classifier weights (forming Equiangular Tight Frame (ETF)) for each object class to induce the maximal separable classifier even under the class imbalance setting. The pre-fixed classifier weights $\mathbf{W}_{ETF}$ are given as:
\begin{equation}
    \mathbf{W}_{ETF} = \sqrt{\frac{\mathcal{C}_{o}}{\mathcal{C}_{o}-1}} \mathbf{U} \left( \mathbf{I}_{\mathcal{C}_{o}} - \frac{1}{\mathcal{C}_{o}} \mathbf{1}_{\mathcal{C}_{o}} \mathbf{1}_{\mathcal{C}_{o}}^T\right)
\vspace{-1mm}
\end{equation}
 where $\mathbf{W}_{ETF} = [\mathbf{w}_1, \mathbf{w}_2,..... \mathbf{w}_{\mathcal{C}_{o}}] \in R^{d \times \mathcal{C}_{o}}$, $\mathbf{U} \in R^{d \times \mathcal{C}_{o}}$, satisfies $\mathbf{U}^T \mathbf{U} = \mathbf{I}_{\mathcal{C}_{o}}, \mathbf{I}_{\mathcal{C}_{o}}$ is the identity matrix, and $\mathbf{1}_{\mathcal{C}_{o}}$ is an all-ones vector. The object classification loss:
 \begin{equation}
     \mathcal{L}_o(x_{o_i}, \mathbf{W}_{ETF}) = \frac{1}{2}\left( \mathbf{w}_{c_{o_i}}^T \hat{x}_{o_i} - 1\right)
\vspace{-2mm}
 \end{equation}
where $\hat{x}_{o_i} = x_{o_i} / ||x_{o_i}|| \textrm{ and } x_{o_i} \in X_{out}^{(n)}$ and $\mathbf{w}_{c_{o_i}}$ is the fixed prototype in $\mathbf{W}_{ETF}$ for object class $c_{o_i}$ and we have $||\mathbf{w}_{c_{o_i}}|| = 1$. Finally, the converged features will be aligned with $\mathbf{W}_{ETF}$, and thus the ETF structure instructed by neural collapse is attained. The theoretical construction of the loss has been proved in \cite{yang2023neural}.
\subsection{Correlation-Aware Predicate Embedding}
\label{coreT}
The relationship between objects is governed by three types of correlations: a) \textit{spatial correlation between predicates}, b) \textit{temporal correlation between predicates}, and c) \textit{predicate-object correlation across the video frames}. We propose to model these correlations using the Vanilla Transformer \cite{vaswani2017attention}. For each object $o_i$ of predicted class $c_{o_i}$ obtained from object detection (Section \ref{objd}), the input to the transformer encoder is the set of features describing the relation with each object $o_j$ detected in all the frames where $o_i$ is detected. The input is constructed as follows:
\begin{equation}
    r_{i,j}^t =  concat(\mathbf{x}_{c_{o_i}}, f_u(\mathbf{u}_{ij}^t + f_{box}(\mathbf{b}_i^t, \mathbf{b}_j^t)), f_I(t)) 
\end{equation}
where $\mathbf{x}_{c_{o_i}} \in X_{out}^{(n)}$ is the feature representation of object $o_i$ belonging to class $c_{o_i} \in [1, \mathcal{C}_o] $, $u_{ij}^t \in \mathbb{R}^{256 \times 7 \times 7} $ is the feature map of the union box computed by RoIAlign \cite{he2017mask}. $f_u, f_I$ is the FFN based on non-linear projections, and $f_{box}$ is the bounding box to feature map projection of \cite{zellers2018neural}. Both are configured to produce d-dimensional relation features. $f_I$ serves as positional embeddings denoting the frame index. The single encoder input consists of both spatial and temporal relation features between object $o_i$ and all other objects ${o_1, o_2,....o_j}$. Specifically, it is denoted as $R_{i} = {r_{i,1}^{t_1}, r_{i,2}^{t_2},.....r_{i,j}^{t_j}}$, where $t_j$ are the frame indices in which $o_i$ and $o_j$ are detected simultaneously. The transformer decoder leverages masked self-attention, and its input is a set of object representations, $\{\mathbf{x}_{c_{o_1}}, \mathbf{x}_{c_{o_2}}..., \mathbf{x}_{c_{o_j}}\}$, corresponding to objects $\{o_1, o_2,.....o_j\}$, detected in all the frames where $o_i$ is detected. Since the input to the transformer encoder contains all predicate (relation) features across the frames, and hence, with multi-head self-attention, it models both spatial and temporal correlations between predicates. Similarly, the cross-attention between the encoder and decoder models the predicate-object correlation. The predicate embeddings at the output of the transformer decoder are denoted as $\hat{r}_{tem}^k = \{\hat{r}_{i,j}^t\} \forall k \in [1, N(t)], t \in [1, T]$. At transformer decoder, we use the sliding window of size 10 for predicate representation with relating objects. 
\vspace{-1mm}
\subsection{Debiased Predicate Embedding}
\label{debar}
\vspace{-1mm}
 Since the relations between objects are highly imbalanced, the representation becomes biased towards popular ones. Therefore, to produce unbiased relation embeddings, we propose updating correlation matrices as a weighted average of the correlation matrices across the epochs, where the weight is determined by the decay factor. For each triplet $\{o_i, r_{i,j}, o_j\}$ the softmax attention scores between the transformer encoder and decoder across all the layers are stored. As some predicates are rare and thus more prone to bias, updating their correlation using a running average will generate debiased embeddings. Let's denote the running average as the stored correlation matrix between every object pair for all relations as $\mathcal{M}_{e-1}$ at the end of the previous epoch $e-1$, and the attention matrix (from Transformer cross-attention) at the current ongoing epoch $e$ as $A_e$. During training, we update the attention matrix (denoted as $\hat{A}_e$) using decaying factor $\eta$ as the training progresses, given as:
\begin{multline}
        \hat{A}_{e}(o_i, r_{i,j}, o_j) = \eta * A_{e}(o_i, r_{i,j}, o_j) + \\ (1 - \eta) * \mathcal{M}_{e-1} (o_i, r_{i,j}, o_j)    
\vspace{-4mm}
\label{corup}  
\end{multline}
where $\mathcal{M}_{0} = A_{0}$, $\mathcal{M}_{e} (o_i, r_{i,j}, o_j) =  \hat{A}_{e}(o_i, r_{i,j}, o_j)$ at the end of each epoch. We then use the attention value from $\hat{A}_{e}$ in place of the attention value calculated from $QK^T \in A_e$. This avoids biasing attention weights towards popular predicates. Once the debiased weights are learned during training, we expect that they will generate debiased embeddings during the inference phase. Hence, there is no modification of attention matrices during inference time.
\vspace{-1mm}
\subsection{Predicate Classification}
\vspace{-1mm}
Following \cite{nag2023unbiased}, we adopt classifier framework to handle noisy annotations. Different from \cite{nag2023unbiased}, we propose: 1) an uncertainty-aware mixture of attenuated loss 2) supervised contrastive learning, which incorporates label correlation to improve predicate classification. Specifically, we propose the classifier head as a mixture-of-experts model named mixture logit networks (MLN) and a noise pattern estimation method utilizing the outputs of the MLN. The number of mixtures is denoted as $\mathcal{K}$. \\
\textbf{Uncertainty-Aware Mixture of Attenuated loss}: For a predicate embedding $\mathbf{z}_i$, the aleatoric ($\sigma_a$) and epistemic uncertainty ($\sigma_e$) are computed as:
\begin{align}
        \sigma_e^2 &= \sum_{p=1}^{\mathcal{C}_r}\sum_{k=1}^{\mathcal{K}} \pi_{i,p}^k ||\mu_{i,p}^k - \sum_{j=1}^{\mathcal{K}} \pi_{i,p}^j \mu_{i,p}^j||_{2}^2 \\ 
       \sigma_a^2  &= \sum_{p=1}^{\mathcal{C}_r} \sum_{k=1}^{\mathcal{K}} \pi_{i,p}^k \Sigma_{i,p}^k
\end{align}
where the mean $\mu_{i,p}$, variance $\Sigma_{i,p}^k$, and mixture weights $\pi_{i,p}^k$ are the logits of label $p$ in kth mixture. These are estimated as follows:
\begin{equation}
    \mu_{i}^k = f_{\mu}^{k}(\mathbf{z}_i), \Sigma_{i}^k = \sigma(f_{\Sigma}^{k}(\mathbf{z}_i)), \pi_{i}^k = \frac{e^{f_{\pi}^k(\mathbf{z}_i)}}{\sum_{k=1}^{\mathcal{K}}e^{f_{\pi}^k(\mathbf{z}_i)}}
\end{equation}
where $f_{\mu}, f_{\Sigma}, f_{\pi}$ are the FFN projection functions and $\sigma$ is the sigmoid non-linearity which ensures $\Sigma_{i,p}^k \geq 0$ for the $p^{th}$ predicate class. During training, $\mathbf{z}_i = \hat{r}_{tem}^i$, the mixture of attenuated loss ($\mathcal{L}_{MAL}$) is given as:
\begin{equation}
    \mathcal{L}_{MAL} = \frac{1}{N} \sum_{i=1}^{N} \sum_{p=1}^{\mathcal{C}_r}  \sum_{k=1}^{\mathcal{K}} \pi_{i,p}^k \frac{\mathcal{L}(\mu_{i,p}^k, y_{r_p}^{i})}{\Sigma_{i,p}^k}
\label{MAL}
\end{equation}
where $\mathcal{L}(\mu_{i,p}^k, y_{r_p}^{i})$ is the sigmoidal cross-entropy loss, $y_{r_p}^{i}$ is the ground-truth predicate class mapped to $\mathbf{z}_i$, $\mu_{i,p}^k$ is the logit of label $p$ in the $k^{th}$ mixture. For the corrupted input, it is more likely to make a false prediction; hence, $\Sigma_{i,p}^k$ will increase to reduce the overall loss function for such instances. This, in turn, prevents over-fitting to the corrupted instances, making the model more robust.\\
\textbf{Uncertainty-aware Supervised Contrastive Learning}: The \textsc{MAL} loss function, which independently classifies labels, presents challenges in capturing correlations among co-occurring semantic labels. To address this limitation, we introduce the multi-label contrastive loss, denoted as $\mathcal{L}_{\textrm{MCL}}$. The primary objective of this loss function is to minimize the distance between representations of predicates sharing at least one class with the anchor representation $\hat{r}_{tem}^n$ while maximizing the distance from negative samples that do not share any classes. Let $\mathcal{A}(n) = \{m \in \{N \setminus n\} : \mathcal{Y}_r^n \cdot \mathcal{Y}_r^m \neq 0$, represent the positive set, where $\cdot$ denotes the dot product\}. This set comprises samples that share at least one label with the anchor $\hat{r}_{tem}^n$. Additionally, let $\mathcal{Y}_r(n, m) = \{y_{r_p}^{\{n,m\}} \in \mathcal{Y}_r^{\{n,m\}} \textrm{ s.t. } y_{r_p}^m = y_{r_p}^n = 1\}$ indicate the indices of samples $m$ that possess at least one shared label with $\mathcal{Y}_r^n$. The loss $\mathcal{L}_{\textrm{MCL}}$ is formulated as:
\begin{multline}
\small
        \mathcal{L}_{\textrm{MCL}} = \frac{1}{N}\sum_{n=1}^{n=N} \frac{-1}{|\mathcal{A}(n)|} \sum_{m\in \mathcal{A}(n)} J(n,m) \\ \sum_{y_{r_p} \in \mathcal{Y}_r(n, m)}  \left ( \textrm{log} \frac{\textrm{exp}(\rho_{y_{r_p}}^{n,m}/\tau )}{\sum_{i\in N \setminus n} \textrm{exp}(\rho_{y_{r_p}}^{n,i}/\tau )}\right )
\end{multline}
where $J(n,m)$ is the Jaccard similarity between labels $y_{r_p}^m$ and $y_{r_p}^n$, the similarity $\rho_{y_{r_p}}^{n,i}$ is given as: 
\begin{multline}
\small
        \rho_{y_{r_p}}^{n,i} = \left ( \prod_{k=1}^{\mathcal{K}} \left ( \frac{(\Sigma_{n,p}^k)^2 + (\Sigma_{i,p}^k)^2}{2 (\Sigma_{n,p}^k) (\Sigma_{i,p}^k)} \right )^{-\frac{1}{2}} \right ) \\ \textrm{exp}\left ( -\frac{1}{4} \sum_{k=1}^{\mathcal{K}}\frac{(\mu_{n,p}^k - \mu_{i,p}^k)^2}{(\Sigma_{n,p}^k)^2 + (\Sigma_{i,p}^k)^2}\right ) 
\end{multline}
Specifically, $\rho_{y_{r_p}}^{n,i}$ represents the Bhattacharyya coefficient, a commonly used metric for evaluating the similarity between probability distributions in various domains, including computer vision, pattern recognition, and statistical analysis\cite{sinha2020neural, combalia2020uncertainty}.\\
\textbf{EMA Teacher}: During training, we adopt the EMA weight update \cite{khandelwal2023segda, tarvainen2017mean, tranheden2021dacs, araslanov2021self} for transformers in Section \ref{coreT}. Let's say $\phi_T, \theta_T$ are the weights of transformers for teacher and student, respectively. The weight update is then given as:
 \begin{equation}
     \phi_{T,e} = \alpha * \phi_{T,e-1} + (1 - \alpha) * \theta_{T,e}
 \end{equation}
 where $e$ is the training epoch. The EMA teacher effectively an ensemble of student models at different training steps, which is a most widely used learning strategy in semi-supervised setting\cite{french2019semi, hoyer2021three}.
\subsection{Training and Testing}
\textbf{Training}: The entire framework is trained end-to-end minimizing the loss equation:
\begin{equation}
    \mathcal{L} = \mathcal{L}_o + \mathcal{L}_{\textrm{MAL}} + \mathcal{L}_{\textrm{MCL}} - \lambda_1 \sigma_e + \lambda_2 \sigma_a
\end{equation}
where $\lambda_1, \lambda_2$ determines the amount of regularization for aleatoric and epistemic uncertainties respectively. This prevents $\Sigma_{i,p}^k$ to grow indefinitely to minimise $\mathcal{L}_{\textrm{MAL}}$ loss. Further we regularize with $\sigma_e$ to increase epistemic uncertainty, thereby encouraging the utilization of a more mixtures. 

\textbf{Testing}: During testing we utilize the EMA teacher $\phi_T$ to generate the predicate embeddings $\hat{r}_{tem}^i$. The debiasing of predicate embeddings is only limited to training. These predicate embeddings are then passed to MLN which outputs the predicate confidence scores from all mixture components. The combined predicate confidence score $\hat{y}_{r_p}^{i}$ from $\mathcal{K}$ mixtures is calculated as:
\begin{equation}
\hat{y}_{r_p}^{i} = \sum_{k=1}^{\mathcal{K}} \pi_{i,p}^k \frac{\mu_{i,p}^k}{\Sigma_{i,p}^k}
\end{equation}
\section{Experiments}
\label{sec:experiments}
\textbf{Dataset}: Following previous works \cite{nag2023unbiased, cong2021spatial}, we also evaluated our method on the most widely used benchmark, Action Genome \cite{ji2020action}. Action Genome is the largest benchmark for video SGG; it is built on top of Charades\cite{sigurdsson2016hollywood}. It contains 476,229 bounding boxes for 35 object classes (without person) and 1,715,568 instances of 26 predicate classes annotated for 234,253 frames. For all experiments, we use the same training and test split following \cite{nag2023unbiased, cong2021spatial, li2022dynamic}.\\
\textbf{Metrics and Evaluation Setup}: We evaluated the performance of \textsc{FloCoDe} with popular metrics, namely, recall@K (i.e., R@K) and mean-recall@K (i.e., mR@K), for $K = [10, 20, 50]$. R@K measures the ratio of correct instances among the top-K predicted instances with the highest confidence, but this is biased towards frequent predicate classes \cite{tang2019learning}, whereas mR@K averages out the R@K over all relationships. Hence, mR@K is a more reliable metric for balanced evaluation across predicates \cite{tang2019learning}.\\
\textbf{Tasks}: Following previous works \cite{cong2021spatial, ji2020action, krishna2017visual, teng2021target}, we also evaluated our method on three different experimental tasks: 1) \textbf{Predicate Classification} (\textit{PREDCLS}): predict the predicate class of object pairs, given the ground-truth bounding boxes and labels of objects. 2) \textbf{Scene graph classification} (\textit{SGCLS}): predict both predicate labels and the category labels of objects given the bounding boxes of objects. 3) \textbf{Scene graph detection} (\textit{SGDET}): simultaneously detects objects appearing in the frame and the predicate labels of each object pair in a frame. Following, we also evaluated our method using two setups: a) \textbf{With Constraint} and b) \textbf{No Constraints}. Later one allows each object pair to have more than one predicates simultaneously while the former one restricts to only one predicate.\\
\begin{table*}[!htb]
\centering
\caption{Comparative results for SGDET task, on AG\cite{ji2020action} in terms of mean-Recall@K and Recall@K, best results are in bold. }
\resizebox{\textwidth}{!}{%
\begin{tabular}{lcccccccccccc}
\toprule
\multirow{2}{*}{\textbf{Method}} & \multicolumn{6}{c}{\textbf{With Constraint}}                                                     & \multicolumn{6}{c}{\textbf{No Constraint}}                                                       \\ \cmidrule(l{2pt}r{2pt}){2-7} \cmidrule(l{2pt}r{2pt}){8-13} 
                                 & \textbf{mR@10} & \textbf{mR@20} & \textbf{mR@50} & \textbf{R@10} & \textbf{R@20} & \textbf{R@50} & \textbf{mR@10} & \textbf{mR@20} & \textbf{mR@50} & \textbf{R@10} & \textbf{R@20} & \textbf{R@50} \\ \midrule \midrule
RelDN                            & 3.3            & 3.3            & 3.3            & 9.1           & 9.1           & 9.1           & 7.5            & 18.8           & 33.7           & 13.6          & 23.0          & 36.6          \\
HCRD supervised                  & -              & 8.3            & 9.1            & -             & 27.9          & 30.4          &  -              &     -           &     -           &     -          &      -         &    -           \\
TRACE                            & 8.2            & 8.2            & 8.2            & 13.9          & 14.5          & 14.5          & 22.8           & 31.3           & 41.8           & 26.5          & 35.6          & 45.3          \\
ISGG                             & -              & 19.7           & 22.9           & -             & 29.2          & 35.3          &     -           &     -           &   -             &    -           &     -          &       -        \\
STTran                           & 16.6           & 20.8           & 22.2           & 25.2          & 34.1          & 37.0          & 20.9           & 29.7           & 39.2           & 24.6          & 36.2          & 48.8          \\
STTran-TPI                       & 15.6           & 20.2           & 21.8           & 26.2          & 34.6          & 37.4          &    -            &       -         &      -          &        -       &       -        &        -       \\
APT                              & -              & -              & -              & 26.3          & 36.1          & 38.3          &       -         &       -         &     -           & 25.7          & 37.9          & 50.1          \\
TEMPURA                          & 18.5           & 22.6           & 23.7           & 28.1          & 33.4          & 34.9          & 24.7           & 33.9           & 43.7           & 29.8          & 38.1          & 46.4          \\
FloCoDe                          &  \textbf{22.6}             &   \textbf{24.2}            &  \textbf{27.9}             &    \textbf{31.5}            &    \textbf{38.4}           &    \textbf{42.4}           &      \textbf{28.6}          &      \textbf{35.4}          &      \textbf{47.2}         &     \textbf{32.6}          &    \textbf{43.9}           &    \textbf{51.6}           \\ \bottomrule
\end{tabular}%
}
\label{with_no_sgdet}
\vspace{-3mm}
\end{table*}
\begin{table*}[!htb]
\centering
\caption{Comparative results for PREDCLS and SGCLS task, on AG\cite{ji2020action} in terms of mean-Recall@K, best results are in bold.}
\resizebox{\textwidth}{!}{%
\begin{tabular}{lcccccccccccc}
\toprule
\multirow{3}{*}{\textbf{Method}} & \multicolumn{6}{c}{\textbf{With Constraint}}                                                                                                                          & \multicolumn{6}{c}{\textbf{No Constraints}}                                                                                                                                              \\ \cmidrule(l{2pt}r{2pt}){2-7}\cmidrule(l{2pt}r{2pt}){8-13} 
                                 & \multicolumn{3}{c}{PredCLS}                                                       & \multicolumn{3}{c}{SGCLS}                                                         & \multicolumn{3}{c}{PredCLS}                                                       & \multicolumn{3}{c}{SGCLS}                                                         \\ \cmidrule(l{2pt}r{2pt}){2-4} \cmidrule(l{2pt}r{2pt}){5-7} \cmidrule(l{2pt}r{2pt}){8-10} \cmidrule(l{2pt}r{2pt}){11-13}    
                                 & \multicolumn{1}{l}{mR@10} & \multicolumn{1}{l}{mR@20} & \multicolumn{1}{l}{mR@50} & \multicolumn{1}{l}{mR@10} & \multicolumn{1}{l}{mR@20} & \multicolumn{1}{l}{mR@50} & \multicolumn{1}{l}{mR@10} & \multicolumn{1}{l}{mR@20} & \multicolumn{1}{l}{mR@50} & \multicolumn{1}{l}{mR@10} & \multicolumn{1}{l}{mR@20} & \multicolumn{1}{l}{mR@50} \\ \midrule \midrule
RelDN                            & 6.2                       & 6.2                       & 6.2                       & 3.4                       & 3.4                       & 3.4                       & 31.2                      & 63.1                      & 75.5                      & 18.6                      & 36.9                      & 42.6                      \\
TRACE                            & 15.2                      & 15.2                      & 15.2                      & 8.9                       & 8.9                       & 8.9                       & 50.9                      & 73.6                      & 82.7                      & 31.9                      & 42.7                      & 46.3                      \\
STTran                           & 37.8                      & 40.1                      & 40.2                      & 27.2                      & 28.0                      & 28.0                      & 51.4                      & 67.7                      & 82.7                      & 40.7                      & 50.1                      & 58.8                      \\
STTran-TPI                       & 37.3                      & 40.6                      & 40.6                      & 28.3                      & 29.3                      & 29.3                      & -                         & -                         & -                         & -                         & -                         & -                         \\
TEMPURA                          & 42.9                      & 46.3                      & 46.3                      & 34.0                      & 35.2                      & 35.2                      & 61.5                      & 85.1                      & 98.0                      & 48.3                      & 61.1                      & 66.4                      \\
FloCoDe                          &     \textbf{44.8}                      &  \textbf{49.2}                         &   \textbf{49.3}                        &          \textbf{37.4}                 &      \textbf{39.2}                      &              \textbf{39.4}             &      \textbf{63.2}                     &       \textbf{86.9}                    &     \textbf{98.6}                      &     \textbf{49.7}                      &       \textbf{63.8}                    &        \textbf{69.2}                   \\ \bottomrule
\end{tabular}%
}
\label{with_no_sgcls_predcls}
\vspace{-3mm}
\end{table*}
\begin{table*}[!htb]
\centering
\caption{Comparative results for PREDCLS and SGCLS task, on AG\cite{ji2020action} in terms of Recall@K, best results are in bold.}
\resizebox{\textwidth}{!}{%
\begin{tabular}{lcccccccccccc}
\toprule
\multirow{3}{*}{\textbf{Method}} & \multicolumn{6}{c}{\textbf{With Constraint}}                                                                                                                          & \multicolumn{6}{c}{\textbf{No Constraints}}                                                                                                                           \\ \cmidrule(l{2pt}r{2pt}){2-7}\cmidrule(l{2pt}r{2pt}){8-13} 
                                 & \multicolumn{3}{c}{PredCLS}                                                       & \multicolumn{3}{c}{SGCLS}                                                         & \multicolumn{3}{c}{PredCLS}                                                       & \multicolumn{3}{c}{SGCLS}                                                         \\ \cmidrule(l{2pt}r{2pt}){2-4} \cmidrule(l{2pt}r{2pt}){5-7} \cmidrule(l{2pt}r{2pt}){8-10} \cmidrule(l{2pt}r{2pt}){11-13} 
                                 & \multicolumn{1}{l}{R@10} & \multicolumn{1}{l}{R@20} & \multicolumn{1}{l}{R@50} & \multicolumn{1}{l}{R@10} & \multicolumn{1}{l}{R@20} & \multicolumn{1}{l}{R@50} & \multicolumn{1}{l}{R@10} & \multicolumn{1}{l}{R@20} & \multicolumn{1}{l}{R@50} & \multicolumn{1}{l}{R@10} & \multicolumn{1}{l}{R@20} & \multicolumn{1}{l}{R@50} \\ \midrule \midrule
RelDN                            & 20.3                      & 20.3                      & 20.3                      & 11.0                      & 11.0                      & 11.0                      & 44.2                      & 75.4                      & 89.2                      & 25.0                      & 41.9                      & 47.9                      \\
TRACE                            & 27.5                      & 27.5                      & 27.5                      & 14.8                      & 14.8                      & 14.8                      & 72.6                      & 91.6                      & 96.4                      & 37.1                      & 46.7                      & 50.5                      \\
STTran                           & 68.6                      & 71.8                      & 71.8                      & 46.4                      & 47.5                      & 47.5                      & 77.9                      & 94.2                      & 99.1                      & 54.0                      & 63.7                      & 66.4                      \\
STTran-TPI                       & 69.7                      & 72.6                      & 72.6                      & 47.2                      & 48.3                      & 48.3                      & -                         & -                         & -                         & -                         & -                         & -                         \\
APT                              & 69.4                      & 73.8                      & 73.8                      & 47.2                      & 48.9                      & 48.9                      & 78.5                      & 95.1                      & 99.2                      & 55.1                      & 65.1                      & 68.7                      \\
TEMPURA                          & 68.8                      & 71.5                      & 71.5                      & 47.2                      & 48.3                      & 48.3                      & 80.4                      & 94.2                      & 99.4                      & 56.3                      & 64.7                      & 67.9                      \\
FloCoDe                          &    \textbf{70.1}                        &                \textbf{74.2}           &       \textbf{74.2}                    &         \textbf{48.4}                  &      \textbf{51.2}                     &           \textbf{51.2}                &          \textbf{82.8}                 &  \textbf{97.2}                         &         \textbf{99.9}                  &      \textbf{57.4}                     &    \textbf{66.2}                        &    \textbf{68.8}                       \\ \bottomrule
\end{tabular}%
}
\label{with_no_r_pred_sg_cls}
\vspace{-3mm}
\end{table*}
\textbf{Implementation details}: Following previous works \cite{cong2021spatial, ji2020action, krishna2017visual, teng2021target, nag2023unbiased}, we adopted Faster R-CNN \cite{ren2015faster} with a ResNet-101 \cite{he2016deep} backbone as the object detector. The object detector was trained on the training set of Action Genome \cite{ji2020action}, resulting in a 24.6 mAP at 0.5 IoU with CoCo metrics. To ensure a fair comparison, we utilized this detector across all the baselines. Per-class non-maximal suppression at 0.4 IoU (intersection over union) was applied, following previous works \cite{cong2021spatial, ji2020action, krishna2017visual, teng2021target, nag2023unbiased}, to reduce region proposals provided by the Region Proposal Network (RPN). The parameters of the object detector (excluding the object classifier) remained fixed during training when training scene graph generation models. For correlation-aware predicate embedding, it is necessary to match object pairs across frames. In cases with multiple objects in the same category, we use the IoU between the two objects across different images to match the subject-object pair. The IoU between the bounding box of the object in the previous frame and the object of the same category in the next frame is calculated. If the IoU is higher than 0.8, they are considered the same object. In the event of multiple candidates, the one with the highest IoU is chosen. An AdamW optimizer \cite{loshchilov2017decoupled} is employed with a batch size of 1 and an initial learning rate of $2e^{-5}$. The number of mixture components $\mathcal{K}$ is set to 4 for SGCLS and 6 for PREDCLS and SGDET. The self-attention and cross-attention layers in our framework have 8 heads with $d=1536$, and dropout is set to 0.1. Regularizer hyper-parameters are set as $\lambda_1 = 1, \lambda_2 = 1$. For debiased predicate embedding, initial $\eta$
 is small i.e. 0.1 and we further reduce it with patience of 3. For EMA teacher update, $\alpha = 0.999$ is used. All experiments are conducted on a single NVIDIA RTX-3090.
\begin{figure*}[!htb]
    \centering
    \includegraphics[width=\textwidth]{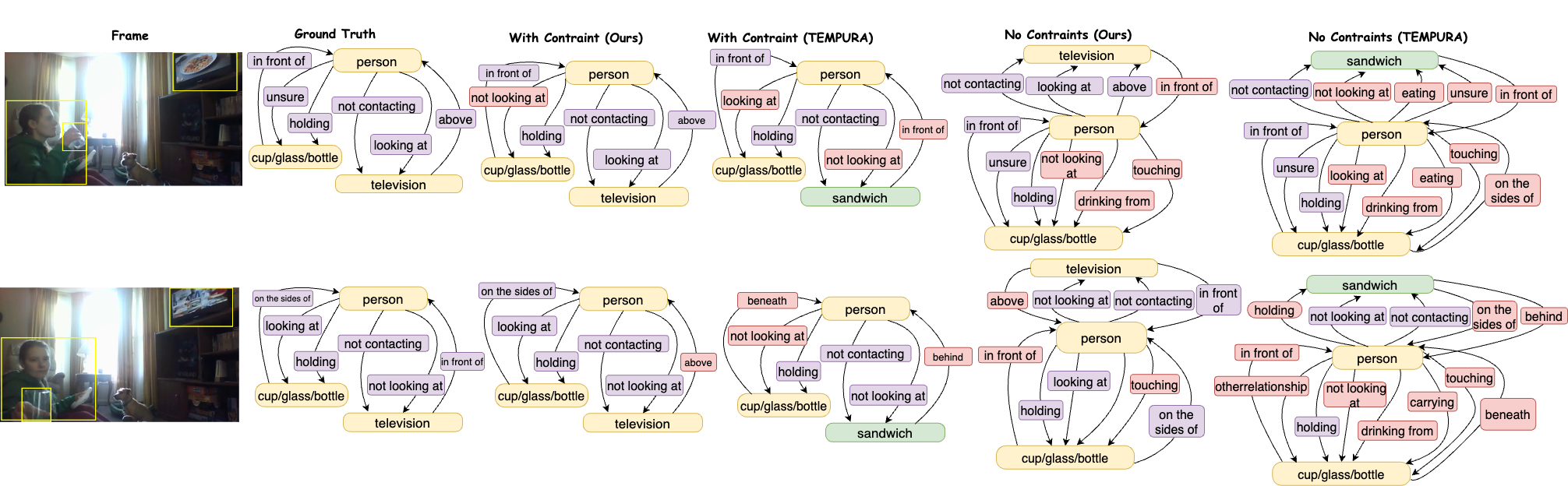}
    \caption{\textbf{Qualitative Comparison} with TEMPURA\cite{nag2023unbiased} for both \textbf{With Constraint} and \textbf{No Constraints} setup. From left to right: input video frames, ground truth graphs, graphs generated by \textsc{FloCoDe}, graphs generated by \textsc{TEMPURA}\cite{nag2023unbiased}.  Incorrect object and predicate predictions are shown in green and red, respectively.}
    \label{qual}
    \vspace{-4mm}
\end{figure*}
\begin{table*}[!htb]
\centering
\caption{\textbf{Ablation Studies}: Importance of \textit{MCL}, \textit{Debiasing}, \textit{TFoD}, \textit{Regularizer} \& \textit{EMA Teacher} for SGCLS and SGDET.}
\resizebox{\textwidth}{!}{%
\begin{tabular}{ccccccccccccc}
\toprule
\multirow{3}{*}{\begin{tabular}[c]{@{}c@{}}Uncertainty-aware\\ Contrastive Learning\end{tabular}} & \multirow{3}{*}{\begin{tabular}[c]{@{}c@{}}Correlation-aware\\ Debiasing\end{tabular}} & \multirow{3}{*}{\begin{tabular}[c]{@{}c@{}}Flow-aware\\ Temporal\\ Consistency\end{tabular}} & \multirow{3}{*}{Regularizer} & \multirow{3}{*}{EMA Teacher} & \multicolumn{4}{c}{With Constraint}                   & \multicolumn{4}{c}{No Constraints}                    \\ \cmidrule(l{2pt}r{2pt}){6-9} \cmidrule(l{2pt}r{2pt}){10-13}
                                                                                                  &                                                                                        &                                                                                              &                              &                              & \multicolumn{2}{c}{SGCLS} & \multicolumn{2}{c}{SGDET} & \multicolumn{2}{c}{SGCLS} & \multicolumn{2}{c}{SGDET} \\ \cmidrule(l{2pt}r{2pt}){6-7} \cmidrule(l{2pt}r{2pt}){8-9} \cmidrule(l{2pt}r{2pt}){10-11} \cmidrule(l{2pt}r{2pt}){12-13}
                                                                                                  &                                                                                        &                                                                                              &                              &                              & mR@10       & mR@20       & mR@10       & mR@20       & mR@10        & mR@20      & mR@10       & mR@20       \\ \midrule \midrule
-                                                                                                 & -                                                                                      & -                                                                                            &   -                           & -                            &      27.2       &     28.0        &    16.5           &     20.8         &  40.7             &    50.1        &      20.9         &     29.7        \\ \midrule
                                                                                 \checkmark                 & -                                                                                      &  \checkmark                                                                                            &    \checkmark                          &    \checkmark                          &  34.1           &   33.8          &  19.6           &  22.1           &  46.9            &  61.1          &   26.6          &  32.7           \\
-                                                                                                 &           \checkmark                                                                             &                                                        \checkmark                                      &      \checkmark                        &          \checkmark                    &     33.6        &    34.3        &    19.4         &   21.8          &   46.2           &   60.6         &    25.8         &    32.5         \\
                                                                         \checkmark                         &       \checkmark                                                                                 & -                                                                                            &                        \checkmark      &        \checkmark                      &  32.2          &  33.4           &   18.1          &    19.8         &    45.9          &     59.1       &   21.8          &   31.6          \\
                                                                         \checkmark                         &              \checkmark                                                                          &      \checkmark                                                                                        & -                            &     \checkmark                         &  35.8           &    36.6         &    21.2         &     22.7        &  48.3            &    61.4        &  27.5           &   34.4          \\
                                                                                           \checkmark       &              \checkmark                                                                          &    \checkmark                                                                                          &      \checkmark                        & -                            &     36.7        &   38.8          &     22.1        &   23.8          &    49.2          &   62.9         &   28.3          &  35.2           \\
                                                                                             \checkmark     &             \checkmark                                                                           &    \checkmark                                                                                          &       \checkmark                       &       \checkmark                       &    \textbf{37.4}         &  \textbf{39.2}            &       \textbf{22.6}      &   \textbf{24.2}          &    \textbf{49.7}          &   \textbf{ 63.8 }       &   \textbf{28.6}          &   \textbf{35.4}          \\ \bottomrule
\end{tabular}%
}
\label{ablation}
\end{table*}
\subsection{Comparison with state-of-the-art}
\begin{figure}[!tbh]
  \begin{subfigure}{.33\columnwidth}
  \centering
    \includegraphics[width=\linewidth]{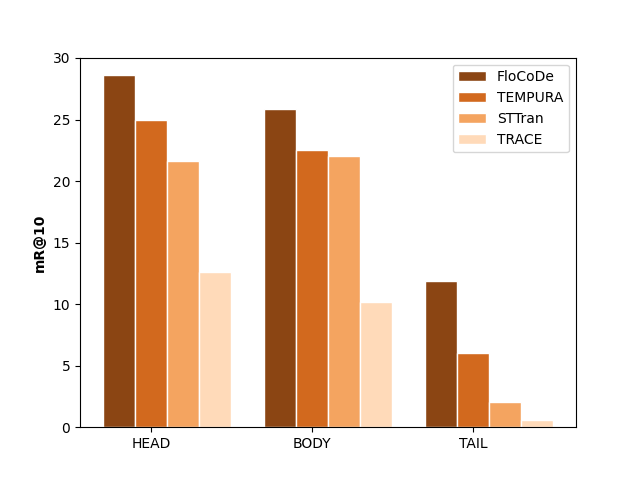}
    \caption{SGDET}
  \end{subfigure}%
  \begin{subfigure}{.33\columnwidth}
  \centering
    \includegraphics[width=\linewidth]{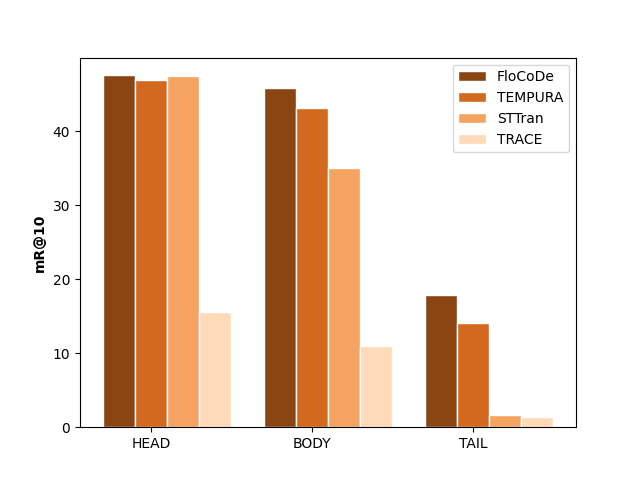}
    \caption{SGCLS}
  \end{subfigure}%
    \begin{subfigure}{.33\columnwidth}
  \centering
    \includegraphics[width=\linewidth]{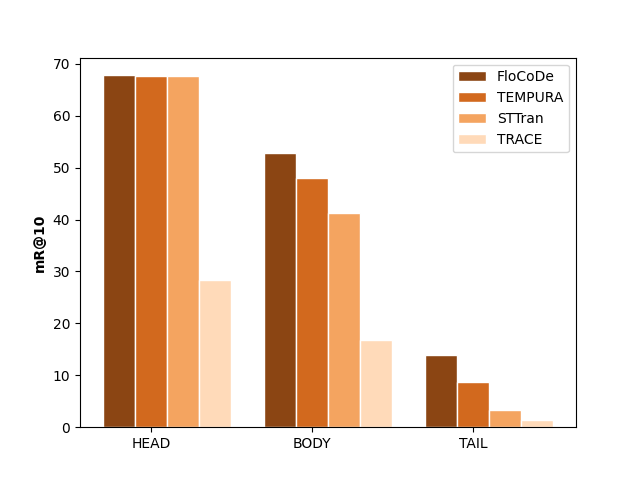}
    \caption{PREDCLS}
  \end{subfigure}
  \newline
   \begin{subfigure}{.33\columnwidth}
  \centering
    \includegraphics[width=\linewidth]{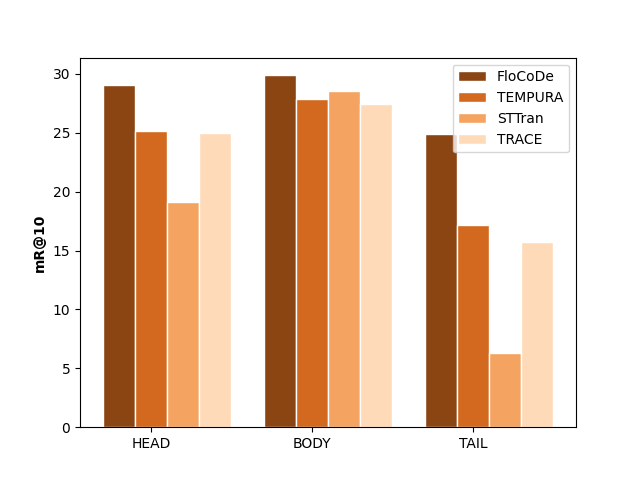}
    \caption{SGDET}
  \end{subfigure}%
  \begin{subfigure}{.33\columnwidth}
  \centering
    \includegraphics[width=\linewidth]{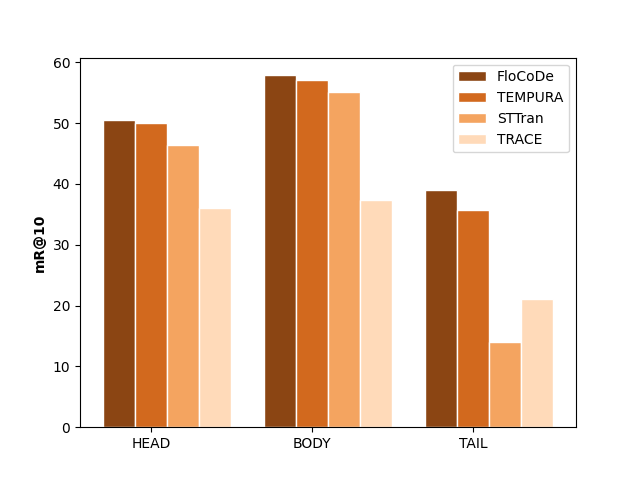}
    \caption{SGCLS}
  \end{subfigure}%
    \begin{subfigure}{.33\columnwidth}
  \centering
    \includegraphics[width=\linewidth]{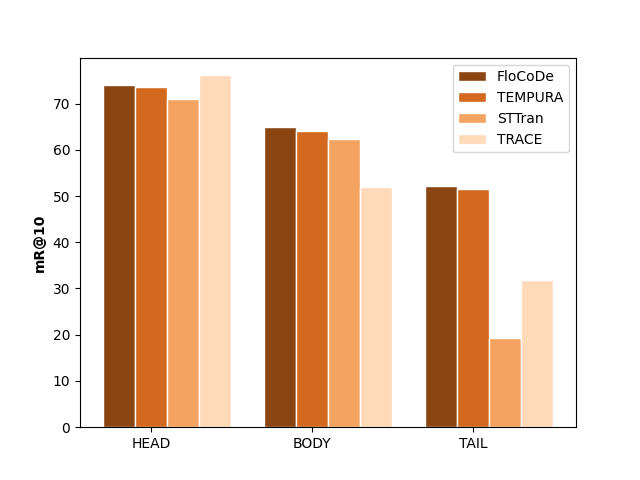}
    \caption{PREDCLS}
  \end{subfigure}
  \caption{Comparison of mR@10 for the HEAD, BODY and TAIL
classes for "with constraint"(top) and "no contraints"(bottom)}
  \label{with_head}
\end{figure}

\begin{figure}
    \centering
\includegraphics[height=5cm,width=0.7\columnwidth]{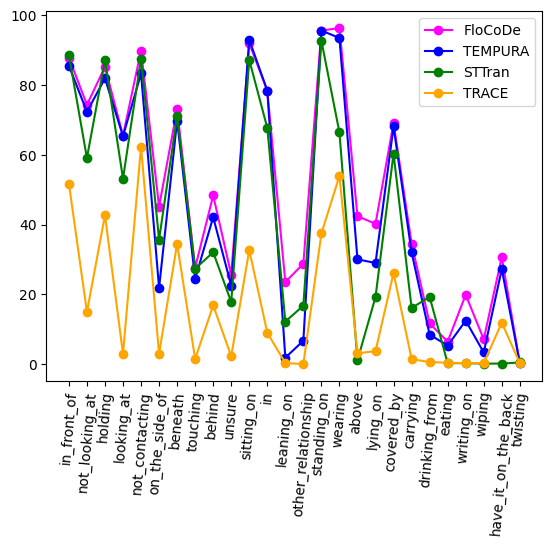}
    \caption{Comparative per class performance for PREDCLS task in R@10 for "with constraint" setup}
    \label{per_class}
\end{figure}

We compared our method \textsc{FloCoDe} with several state-of-the-art methods for dynamic SGG, namely TEMPURA \cite{nag2023unbiased}, STTran \cite{cong2021spatial}, TRACE \cite{teng2021target}, STTran-TPI \cite{wang2022dynamic}, APT \cite{li2022dynamic}, and ISGG \cite{khandelwal2022iterative}. Additionally, we compared our method with ReLDN \cite{zhang2019graphical}, which is a static method. Performance comparisons in terms of mR@K and R@K for \(K = [10, 20, 50]\) are reported in Tables \ref{with_no_sgdet}, \ref{with_no_sgcls_predcls}, and \ref{with_no_r_pred_sg_cls}. These tables contain comparisons with two experimental setups: a) \textbf{With Constraint} and b) \textbf{No Constraints}. For the tasks, i.e., \textit{PREDCLS + SGCLS}, we presented results for these experimental setups, specifically reporting mR@K and R@K in Tables \ref{with_no_sgcls_predcls} and \ref{with_no_r_pred_sg_cls}, respectively. Table \ref{with_no_sgdet} compares results for the task \textit{SGDET}. From the tables, it has been observed that our method consistently outperforms other methods across all tasks and for both experimental setups. Specifically, in comparison to the best baselines, we observe improvements of 4.1\% on \textit{SGDET}-mR@10, 3.4\% on \textit{SGCLS}-mR@10, and 1.9\% on \textit{PREDCLS}-mR@10 under the "\textbf{With Constraint}" setup. Under the "\textbf{No Constraints}" setup, we observe more significant improvements, with 3.9\% on \textit{SGDET}-mR@10, 1.4\% on \textit{SGCLS}-mR@10, and 1.7\% on \textit{PREDCLS}-mR@10. This demonstrates the capability of \textsc{FloCoDe} in generating more unbiased SGG for videos incorporating dynamic fluctuations and long-tailed relations. We further verified this in Figure \ref{with_head} for \textbf{With Constraint} and \textbf{No Constraints}. In these figures, we compared our method on HEAD, BODY, and TAIL classes with mR@10 values. We split the classes into HEAD, BODY, and TAIL with the same definition as mentioned in \cite{nag2023unbiased}. Clearly, \textsc{FloCoDe} improved the performance across all the classes, but the improvement for TAIL classes is more confirming the unbiased predictions. Per-class performance is shown in Fig. \ref{per_class}, comparing with other methods STTran and TRACE, showing improvement at the class level. Additionally, our method outperforms in terms of R@K values, as shown in Table \ref{with_no_r_pred_sg_cls}, demonstrating improvements overall compared to existing methods. This shows that our method has better generalization since it performs better on both mR@K (long-tail) and R@K (overall). Qualitative visualizations are illustrated in Fig. \ref{qual}.
\subsection{Ablation Studies}
\begin{table}[!htb]
\centering
\caption{Results (in mR@10) with varying number of mixtures $\mathcal{K}$ for \textbf{With Constraint} setup}
\begin{tabular}{llllll}
\toprule
\diagbox{Task}{$\mathcal{K}$} & 1 & 2 & 4 & 6 & 8 \\ \midrule
PREDCLS               & 39.8  &  41.2 & 43.4  & \textbf{44.8}  &  44.2 \\
SGCLS                 &  30.2 &  35.5 & \textbf{37.4}  & 36.2  &  35.8 \\
SGDET                 & 16.1  & 18.1  & 21.9  & \textbf{22.6}  & 22.1  \\ \bottomrule
\end{tabular}%
\label{GMM}
\end{table}
We have conducted extensive ablation studies on SGCLS and SGDET tasks. Specifically, we studied the impact of \textit{MCL} (uncertainty-aware contrastive learning), \textit{Debiasing} (correlation-aware debiasing), \textit{TFoD} (flow-aware temporal consistency), \textit{Regularizer} (aleatoric and epistemic regularizer), and \textit{EMA Teacher}. When all these components have been removed, \textsc{FloCoDe} boils down to the baseline STTran \cite{cong2021spatial}, where the object proposals and predicate embeddings are fed to the FFN layers before finally predicting the predicate class using a classification layer. The results for these ablation studies are presented in Table \ref{ablation}. \textbf{Uncertainty Attenuation and Debiasing}: We first discuss the impact of uncertainty-aware contrastive learning and correlation-aware debiasing. First, we remove the loss $\mathcal{L}_{\textsc{MCL}}$ to study the improvement on top of $\mathcal{L}_{\textsc{MAL}}$. In the second case, we remove the correlation-aware debiasing during training. The results for these are shown in Table \ref{ablation}, rows 1 and 2. Comparing the resulting models with \textsc{FloCoDe} shows a significant drop in mR@10 values, indicating the value addition from each of these in generating unbiased SGG. This also shows that both can address the bias, specifically, contrastive learning deals with label correlation especially for long tailed classes and hence focus to generate unbiased predicate embeddings. \textbf{Temporally Consistent Object Detection}: Table \ref{ablation} (row 3) clearly illustrates the impact of flow-aware detection on ensuring temporal consistency in object identification. The absence of \textit{TFoD} leads to a noticeable performance decline when contrasted with \textsc{FloCoDe}, underscoring the pivotal role of accurate object detection as a key bottleneck in SGG methods. For the \textit{PREDCLS} task, leveraging only ground-truth boxes and labels results in significantly higher mR@k and R@K values compared to other tasks. Regarding the \textbf{Uncertainty Regularizer and EMA Teacher} components (rows 4 and 5), their ablation underscores their crucial role in reducing noise, especially for TAIL classes. Aleatoric and epistemic regularization helps better prediction of the noise distribution i.e. mixture variances. EMA teachers contribute to balanced predicate embeddings, enhancing performance specific to their respective classes. \textbf{Number of Mixtures} $\mathcal{K}$: The performance of \textsc{FloCoDe} with varying numbers of mixtures in MLN is shown in Table \ref{GMM}. The number of mixtures between 4 to 6 is optimal.
\vspace{-2mm}
\section{Conclusion}
\label{sec:conclusion}
In summary, this paper proposes \textsc{FloCoDe}, a method for generating unbiased dynamic scene graphs from videos. By addressing issues such as biased scene graph generation and the long-tailed distribution of visual relationships, \textsc{FloCoDe} achieves significant performance gains. Through features like flow-aware temporal consistency, correlation debiasing, label correlation and uncertainty attenuation it offers a robust solution for capturing accurate scene representations in dynamic environments.
{
    \small
    \bibliographystyle{ieeenat_fullname}
    \bibliography{main}
}
\end{document}